\documentclass[12pt, a4paper]{article}

\usepackage[margin=1in]{geometry}  
\usepackage{graphicx}  
\usepackage{amsmath}   
\usepackage{amssymb}   

\begin{document}

\title{Elephants and Algorithms: A Review of the Current and Future Role of AI in Elephant Monitoring}

\author{
    Leandra Brickson\thanks{Colossal Biosciences, Dallas, TX, USA} 
 \thanks{publications@colossal.com}\\
    \and
    Libby Zhang\footnotemark[1] \\
    \and
    Fritz Vollrath\thanks{University of Oxford, Oxford, United Kingdom} 
 \thanks{Save the Elephants, Nairobi, Kenya} \\
    \and
    Iain Douglas-Hamilton\footnotemark[4] \\
    \and
    Alexander J. Titus\footnotemark[1] \thanks{International Computer Science Institute, Berkeley, CA, USA} \\
}

\date{}
\maketitle

\begin{abstract}
Artificial intelligence (AI) and machine learning (ML) present revolutionary opportunities to enhance our understanding of animal behavior and conservation strategies. Using elephants, a crucial species in Africa and Asia's protected areas, as our focal point, we delve into the role of AI and ML in their conservation. Given the increasing amounts of data gathered from a variety of sensors like cameras, microphones, geophones, drones, and satellites, the challenge lies in managing and interpreting this vast data. New AI and ML techniques offer solutions to streamline this process, helping us extract vital information that might otherwise be overlooked. This paper focuses on the different AI-driven monitoring methods and their potential for improving elephant conservation. Collaborative efforts between AI experts and ecological researchers are essential in leveraging these innovative technologies for enhanced wildlife conservation, setting a precedent for numerous other species.
\end{abstract}

\section{Introduction}

Elephants hold significant intrinsic, biological, ecological, and human cultural value. They possess unique genetic and physiological features \cite{Dagenais2021}, exhibit high-levels of individual cognitive and emotional intelligence \cite{Chevalier-Skolnikoff1993, Plotnik2006, Goldenberg2020}, maintain complex social behaviors and structures \cite{Mccomb2000, Mccomb2003, Wittemyer2005, DeSilva2011, Lee2014}, serve as keystone species in their respective ecosystems \cite{Santiapillai2004, Haynes2012, Guldemond2017}, are highly valued by their local communities \cite{Kahindi2001} and serve as flagship species for animal conservation \cite{Bowen2002}.

Unfortunately, the three extant species of elephants, African Savannah elephants (\textit{Loxodonta africana}), African forest elephants (\textit{L. cyclotis}), and Asian forest elephants (\textit{Elephas maximus}), face multiple continued threats to their survival including habitat loss \cite{Santiapillai1997,Saaban2011}, human-elephant conflict \cite{Perera2009}, and poaching \cite{Blake2007, Whittemyer2015}. This paper outlines how conservationists can leverage advances in the fields of artificial intelligence and machine learning to develop new solutions and capabilities for studying, monitoring, and managing elephants and their habitats to ensure their survival.

Elephant conservation challenges have grown more acute, varying across different elephant populations. It's essential to note that while all elephant populations face a multitude of problems, certain challenges have become predominant for specific populations. Forest elephants, especially from the Congo basin, face severe threats from poaching due to the allure of their hard ivory \cite{Blake2007, Whittemyer2015}. Savannah elephants are grappling with habitat fragmentation caused by human expansion, disrupting their traditional migration routes \cite{Gross2022, Di2021, Hoare2000, Sach2019} and exacerbating human-elephant conflicts particularly near farmland borders \cite{Perera2009, Wenborn2022, Gross2022}. Meanwhile, Asian elephants, heavily impacted by habitat fragmentation, suffering from limited genetic diversity due to limited mobility and are frequently compelled to relocate from shrinking habitats, often resulting in heightened conflicts \cite{Santiapillai1997, Saaban2011}. As conservationists continue to tackle these complex issues, it is imperative that they are armed with state of the art tools and monitoring techniques. This includes models and methods for enhancing dataset collection and annotation, particularly in countries where elephant conservation receives limited funding.

Conservation efforts have traditionally relied on a variety of techniques for monitoring elephant populations. These include direct methods, such as aerial and ground surveys \cite{Chase2018,Dunham2015}, as well as indirect methods like analysing dung samples \cite{Barnes1997} and using camera traps \cite{Smit2019}. However, these methods often have limitations. Aerial and ground surveys are labour-intensive, costly, and affected by weather conditions and visibility \cite{Schlossberg2016, Koneff2008, Lamprey2020, Lamprey2020_2}. Dung counts on the other hand, may provide limited information on population demographics and are prone to errors and biases associated with the variation in dung decay rates in different parts of the elephant range. With advances in monitoring technology, wildlife ecology can acquire more informative datasets for conservation efforts with potentially lower cost. As the ease of data collection improves, this consequently leads to higher volumes of data to be collected and analysed.

The fields of artificial intelligence (AI) and machine learning (ML), which deal with high-level semantic reasoning and data-driven pattern recognition, are well-suited to leverage this modern volume of data and to develop new solutions and capabilities for conservation science.
Comprehensive reviews on AI/ML's application in wildlife ecology and conservation can be found in the works of Farley et al. \cite{Farley2018}, Christin et al. \cite{Christin2019}, and Tuia et al. \cite{Tuia2022}. A recurring emphasis of these articles is the need for cross-disciplinary collaboration between conservationists and technologists. Here, we build on these calls by honing in on species-specific conservation priorities, by surveying existing applications of AI techniques and sketching out problem spaces where this class of approaches may be well-suited. This paper is geared towards elephant researchers and conservationists, but it might also be useful for AI/ML practitioners seeking entry points into this field to gain a tangible sense of existing technological hurdles to key elephant conservation goals.

In the following sections, we introduce AI/ML applications by measurement modality: imaging, acoustics, seismic, and molecular. In doing so, we highlight the potential of AI/ML in addressing specific monitoring needs for elephants and discuss current and upcoming AI/ML techniques suitable for conservationists. The final section dives into the broader implications of AI in the field of elephant monitoring, focusing on strategies to reduce costs and to cope with massive new datasets. 
\section{Image and Video Monitoring}
Imaging and video monitoring have been key tools in elephant conservation for a long time. Up until recently, this technology needed a person to analyse the recorded data to annotate elephant sightings, count elephants and record elephant behaviour. However, this field can be greatly enhanced by recent advances in AI monitoring. 

\subsection{Elephant Detection}\label{sec:image:detection}
AI-assisted camera traps such as those partnered with the SMARTParks monitoring system \cite{SmartParks}, WildEye \cite{WildEye2023}, EarthRanger \cite{EarthRanger} and Mbaza AI \cite{Alim2021} have been increasingly used in parks and protected wildlands for elephant monitoring. These systems employ AI algorithms to detect and identify different species captured by camera traps, thereby helping human researchers sift through and process vast amounts of video data. As a result, they enable identification of elephants entering the camera trap's field of view. Though the technical details of SMART and WildEye systems remain undisclosed, recent publications on species classification using camera trap images report classification accuracies of around 90\% \cite{Tabak2019, Vecvanags2022, Willi2019, Villa2017, Schneider2020}. Mbaza AI, in particular, classifies 25 species with a 96\% accuracy \cite{Alim2021}. While the performance of these systems is impressive, they have been reported to struggle when transferring to new environments \cite{Schneider2020} and the limited visible range of camera traps constrains the coverage of this as a census technique.

AI techniques have been integrated with aerial surveys for enhanced elephant monitoring. Cameras mounted on survey aircraft capture images, and AI algorithms are used to count and track elephants in the recorded data \cite{Lamprey2020,Gonzalez2016,WildMe2020}. Often these tools are used to detect the sparse appearance of elephants within a dataset of hundreds of thousands of images. AI detection techniques eliminate the need for human presence on the aircraft, the manual inspection of the images after capture, and reduces the chances of miscounting due to human error and fatigue.

Recently, AI techniques have been utilised to count and monitor habitat use of elephants from high-resolution satellite images \cite{Duporge2021}. This approach shows great promise for wildlife monitoring, particularly in areas where elephants are not obstructed under canopies and sparsely dispersed across large home ranges such as desert environments. However, several technical challenges limit the widespread adoption of this method. These challenges include the high cost of acquiring satellite images, the lack of high-resolution satellite coverage, and the inability to detect obstructed elephants under canopy. Although the increase of high-resolution satellites and improvement of AI techniques for elephant detection may make this approach more feasible in the future, the authors suggest that oblique camera count is currently more practical for detecting elephants, as it provides higher resolution images and allows for detection of elephants under tree canopies.

Finally, in the context of human-elephant conflicts (HECs), elephant detection techniques using camera traps have been employed to detect elephants approaching farmlands, serving as a warning to farmers about impending crop raids \cite{Premarathna2020}. When these sensors activate, local farmers are informed and in some cases the elephants can be driven away.

\subsection{Individual Identification}
Deep learning has achieved promising initial results on individual elephant identification, however higher accuracy is needed before these methods are practical in the field. Korschens et al. \cite{Korschens2018} trained a Convolutional Neural Network (CNN) to recognise individual elephants from a group of 276 with a 74\% accuracy, using only a few training images per individual. Other studies have trained various neural network architectures to examine images or the contours of elephant ears for individual recognition, with the best results yielding an 88\% accuracy \cite{De2022, Weideman2020}. Recent work has combined an ensemble of current elephant identification techniques into a tool called ElephantBook \cite{Kulits2021}, with this ensemble, the system achieved 93\% accuracy in correctly identifying individuals within the top 15 matches.

Improving the accuracy and robustness of individual identification from images could potentially improve elephant monitoring and can be used to track individual elephants across extensive time-frames, and be used to recognise repeat offenders of human-elephant conflicts \cite{Srinivasaiah2019}.
Machine learning methods to identify individuals have been demonstrated in small animals using whole-body morphology such as fish \cite{Ditria2020}) or in patterned animals such as cheetas, tigers, and giraffes \cite{cheema2017patternid,shi2020tigerstripes}. However, the sheer size of elephants makes it difficult to capture their whole body, particularly in forested settings where visual obstruction due to dense vegetation requires cameras to be placed close to the anticipated sites of anticipated elephant crossing. Additionally, elephants lack consistent distinguishing patterning variation; instead, expert reviewers typically use morphological features based on ears (such as folds, lope shapes, and tears or holes), tusk or tush characteristics (such as angle and symmetry), tail length and brush type, and back slope to identify adult elephants \cite{foley2010tanzania,Srinivasaiah2019,Montero2023}.
Focused development of methods to characterise these morphological features in concert would greatly improve automatic identification of individuals. Techniques from AI-based human facial recognition techniques \cite{Taskiran2020} can also be adapted to body morphology recognition. These techniques have been used with success in bears \cite{Clapham2020,Clapham2022}, primates \cite{Shukla2019} and giant pandas \cite{Hou2020} for individual face recognition. 

\subsection{Automated Behaviour Analysis}\label{behavior}

An emerging area with important conservation impacts is the study of individual elephant behaviour and its application in detecting perceived threats by elephants, understanding individual and group decision-making processes, and mitigating human-elephant conflicts.
Direct field observations \cite{Kahl2002, DouglasHamilton2006} and long-term ethological studies \cite{Turkalo2013,Moss2011} continue to provide the foundation for understanding the rich and complex behaviours exhibited by elephants. These studies have built up invaluable collective knowledge of the form, function, and contexts of elephant behaviour, and have accumulated into living multimedia resources such as the Elephant Ethogram book \cite{Poole2021}.
Maturation of machine vision methods, coupled with advances in remote and sustained observation of elephant behaviour, offer a way to extend this knowledge and quantify individual behaviour at greater scales through automated behaviour analysis (ABA).

\subsubsection{Behaviour observation and representation}

Automated learning of visual-based behaviours analyse body posture representations over time. As such, this approach is dependent on numerous other technically challenging tasks, namely video acquisition that balances spatial resolution with field of view, followed by robust postural estimation. ABA methods assume that the significant portion of variability in its input data is due to behavioural variability and not due to imaging or environmental variability; thus, it is equally as important to understand the capabilities and existing challenges of these preceding technical components. 

Video-camera traps and camera-equipped unmanned aerial vehicles (UAVs) are offering researchers and conservationists more sustained and previously inaccessible views of elephant behaviour at high spatiotemporal resolution. Video-camera traps, which remain vital sensors for forest-dwelling elephants and cause minimal behavioural disruption, have been used successfully to classify behaviour by manual human review \cite{vandeWater2020}.

These systems are in the same technology class as camera traps and thus share similar methodological and logistical considerations, including camera placement, sensor sensitivity and resolution, and environmental resilience \cite{OConnell2011}.

The review by Dell et al. \cite{Dell2014} on the challenges of image-based tracking in the field and their call to developers remain relevant even in light of the rapid progress in machine vision that has occurred in subsequent years.

One significant advance has been that of automatic pose estimation. Postural representations capture the relevant features of an animal’s pose, such as appendage configuration using keypoint-based representations or soft-tissue and body condition using dense point cloud or surface mesh representations.
We refer readers to Mathis et al. \cite{Mathis2020} and Jiang et al. \cite{Jiang2022} for a review of animal pose estimation.

A persistent challenge is the standardisation of pose and robustness to visual occlusion, irrespective of an animal’s distance or orientation to the camera. This is an inherent challenge in observing with a single camera and representing pose in two-dimensions (2D), and can lead to corrupted downstream behavioural inferences that would otherwise be resolved with three-dimensional (3D) pose representations \cite{gunel2019deepfly3d}.

Multiple cameras may be used to resolve this inherent depth ambiguity, but this introduces additional data synchronisation challenges and increases system cost. Research in machine learning methods for single-view (or monocular) 3D pose estimation provides a promising alternative solution \cite{liu2022monocular3d}. The most feasible of these approaches is the use of shape and skeletal priors, whose utility have been demonstrated by existing methods in 3D animal pose estimation \cite{zuffi2017smal, Zhang2021, hu2023monocular3dmouse}.

UAVs provide unprecedented field-of-views of animal groups and mobility in following them over difficult terrain in open landscapes such as the African savanna \cite{Schad2023,Koger2023}. We note that overhead observation, as will be discussed, is limited in its ability to capture elephant behaviour occuring under trees, and even thermal imaging cameras do not have enough contrast or resolution during the daytime due to ambient temperatures.

UAVs collecting oblique video footage overcome the fixed nature of ground-based video camera traps and can be flexibly positioned to control the level of observation detail versus field of view (including at an angle to observe elephants under trees), but they otherwise face many of the same pose estimation challenges described above.

A much simpler imaging condition is that of UAVs collecting overhead video footage, where steady altitude, relatively large distance to subjects, and complete view of gross body position (e.g. orientation and heading) circumvents the need for fuller, 3D descriptions of pose.
The trade-off is naturally in the detail of behavioural observation, but if the focus is on how cumulative behaviour and interactions influence group-level behaviour, then these observations are typically sufficient and already richer than previous sparse and point observations provided by telemetry-based systems \cite{CostaPereira2022Review}.
A key development has been in the robust and high-resolution georeferencing of free-ranging animals and landscape reconstruction from these aerial videos using a combination of data collection protocols, data fusion, and machine vision \cite{Koger2023}. This is particularly relevant for studying elephant behaviour over large areas such as migratory routes or when using multiple drones to capture diffuse group behaviour.
We refer readers to Corcoran et al. \cite{Corcoran2021} and Schad et al. \cite{Schad2023} for more in-depth reviews on the use of drones for studying animal behaviour, and Kroger et al. \cite{Koger2023} for an accessible methodology guide spanning data collection, georeferencing, and individual tracking and pose estimation.
In spite of noise reduction technology, however, one persisting drawback of UAV-based monitoring is the flight noise, which impacts elephant behaviour. The recognised utility of this observational approach has spurred active research by ecologists and conservationists in developing protocols to minimising UAV disturbance and improving habituation of elephants and other wildlife to UAVs \cite{Duporge2021, Hartmann2021,vanVuuren2023}. Future behavioural insights and other conservation insights as a result of aerial-based monitoring may catalyse further hardware development of quieter UAVs for conservation purposes.

\subsubsection{Behaviour analysis}

Machine learning-based behavioural analysis methods have successfully expedited the automated annotation of video data in behavioural assays \cite{Kabra2013, Bohnslav2021, Gabriel2022} and social interactions \cite{nilsson2020simba, Segalin2021, Klibaite2017}, used to discover the structure and dynamics of expressed behaviour \cite{Berman2014, Wiltschko2015,Werkhoven2021} and reveal previously undescribed movement phenotypes in neurodevelopmental models \cite{Wiltschko2015,Klibaite2022}. Collective behaviour analysis has also been applied to many animal species to understand how decisions are made in a group \cite{Couzin2005}. 

This is an active area of research in the fields of neuroscience and behavioural biology, and we refer readers to Anderson et al. \cite{Anderson2014}, Valletta et al. \cite{Valletta2017}, Datta et al. \cite{Datta2019} and Couzin et al. \cite{Couzin2022} for more pedagogical introductions, comprehensive overviews, and nuanced reviews of the field. 

We consider the field of automatic behavioural analysis methods for elephant monitoring in terms of supervised approaches, sometimes referred to automatic behaviour detection or action recognition, and unsupervised approaches, sometimes referred to as deep behavioural phenotyping.
When the detection or study of well-defined behaviours is required, researchers can curate a labelled subset of video frames during which the behaviours of interest is exhibited, and use supervised approaches \cite{Kabra2013,Bohnslav2021,Gabriel2022,Sturman2020}.
Such studies have demonstrated human-level accuracy in detecting whole-body behaviours in mice such as floating, rearing, nose-to-nose in paired individuals, and more localised behaviours such as head dipping, grooming, and scratching \cite{Bohnslav2021,Sturman2020}.
Applying these methods in order to automatically detect heightened attentiveness behaviours in elephants, such as fence touching, ear flapping, and swaying, and quantifying the number of occurrences and duration would be useful in evaluating and sustaining the effectiveness of boundary deterrents such as beehive fences \cite{King2017,vandeWater2020}. 
Automatic behavioural detection methods may also be useful in evaluating proxies of elephant psychological state, such as relaxed versus perceived threat, at key locations such as water sources or wildlife crossings.

When a more extensive set of behaviours, patterns or open-ended ethological questions are of interest, unsupervised deep phenotyping \cite{Berman2014,Wiltschko2015, Klibaite2017,hsu2021bsoid,Luxem2022,Weinreb2023} provides a means to identify behavioural sequences from the statistics of the individual and group postural data alone.

The consistency of discovered sequences with ethologically-relevant behaviour have been experimentally validated via optogenetic stimulation \cite{gunel2019deepfly3d} and pharmaceutical induction \cite{Wiltschko2020}, and demonstrated high enough sensitivity to identify previously undescribed pausing and head-bobbing behavioural phenotypes in genetically-modified mice \cite{Wiltschko2015}.

Deep phenotyping could be used to study the differences in the type and structure of social behaviour between orphaned elephants versus elephants from intact families \cite{goldenberg2017orphaned,stokes2017nocturnal,garai2023translocation}, or translocated versus resident elephants \cite{pinter2009translocation,horner2021introduction}, in order to improve re-introduction and social integration efforts.

We note that, when studying behaviour in natural settings, preference should be given to ABA methods that operate on postural representations, which were previously introduced, rather than single-stage methods that operate directly on pixels.
Single-stage methods, such as \cite{Wiltschko2015,Batty2019,Bohnslav2021}, directly infer behaviour from changes in pixel values and provide a streamlined machine learning pipeline. These methods work well in controlled imaging environments with simple, consistent backgrounds such as laboratory settings, where the changes in pixel values are almost entirely attributable to a behaving animal. However, these methods will suffer when changes in pixel values are due to non-behavioural variables, such as imaging conditions, changing backgrounds, and visual appearance. These challenges are faced even in laboratory settings \cite{gunel2019deepfly3d,hu2023monocular3dmouse}.

On the other hand, if videos are cropped to tightly bound the relevant subjects in order to reduce image-based variability, a few studies have demonstrated high performance using single-stage behavioural inference in more controlled but still complex settings, such as in an enriched, multi-level home cage of group-housed rhesus macaques \cite{Marks2020} and in a natural forest clearing of a resident wild chimpanzees' home range using handheld video recorders \cite{Bain2021}. These demonstrations suggest that single-stage behavioural inference may still find application when fine-tuned for specific camera views and imaging conditions.

To the best of our knowledge, automated vision-based behavioural analysis has not yet been conducted in free-ranging elephants, and only minimally in other wild free-ranging mammals (\cite{Bain2021}: wild chimpanzees, detecting nut cracking and passing food to mouth, \cite{feng2021felines}: wild felines, standing, ambling, and galloping gait detections). A handful of papers have applied automated pose estimation to ground- and aerial-based videos (\cite{wiltshire2023deepwild}: wild apes,\cite{Koger2023}: ungulates), but have not explicitly conducted behavioural analyses.
This may be reflective of the challenge of translating automated behavioural analyses from the lab into the wild, as previously mentioned.
After constraining the variability of ABA inputs, be it pose or pixels, to just behavioural variability, challenges and limitations of existing methods still remain.
Automated behaviour detection faces the standard challenges associated with supervised learning approaches, such as data labelling time, annotator inconsistency, and sampling and temporal bias. The ethological-relevance of the behavioural sequence classes discovered by deep phenotyping require manual human validation and semantic assignment in practice. These also produce more sequence classes than can be distinguishable by eye, and these are typically agglomerated by using a model variant that explicitly accounts for additional sources of behavioural variability, such as speed \cite{costacurta2022twarhmm}, or post-hoc via manual curation \cite{Klibaite2022}.

Additional challenges in modelling individual and group elephant behaviour arise simply because of its complexity and species-specificity, and the development of behavioural methods to capture these nuances has not yet been motivated. For example, elephant interactions can be indirect and take place over the timescale of tens of minutes which laboratory-motivated methods, that are focused on sub-second behavioural precision, are generally not designed to perform inference at these timescales nor have the representational capacity to study these interactions. Moreover, elephants have complex, non-visual modes of signalling, to be explored in future sections, that are still just being understood and which may complicate the inference of indirect interactions within a group. The unprecedented group-level behaviour observations with individual behavioural resolution requires translation of theories from movement ecology \cite{strandburg2015baboons,ozogany2015horses} in order to better understand the group decision-making processes underlying socially hierarchical elephants.

The application of these techniques to behavior patterns relevant to conservation is gradually being realised with advances in machine vision and laboratory-based ABA.
For example, quantifying the "personality" of individual elephants to assess their likelihood of engaging in crop raiding \cite{Hoare1999,Mumby2018,Srinivasaiah2019} would allow the efficient allocation of GPS collars and other resources to monitor high-risk individuals. Remote evaluation of elephant behaviour could draw from studies demonstrated for humans, such as through gait tracking \cite{Han2005,Teepe2021}, and action patterns, such as driving patterns \cite{Enev2016}, to inform wildlife managers about elephant health, wellness, intrinsic states, and evolved behavioural strategies.
Dedicated work by AI/ML practitioners will be required to robustly translate the latest AI and ML methods to continuous, variable observations of elephants in their natural settings, scientific and field experts are needed to guide the behavioural questions and metrics of highest conservation priority, and dedicated investments from interdisciplinary researchers will be necessary to develop novel methods for studying elephant-specific, evolved and emergent behavioural strategies.

\section{Acoustic Monitoring}
Auditory communication plays a pivotal role in the social lives and survival of elephants. Utilizing a diverse repertoire of calls, including infrasonic rumbles, trumpets, and roars, elephants convey essential information related to group cohesion, reproductive status, and alarm signals \cite{Moss2011,Poole2005}. This audio communication contributes significantly to the complex social structure of elephants, facilitating coordination among group members, resource acquisition, and the transmission of knowledge and social learning \cite{Poole2005,Byrne2009}. Consequently, the intricate acoustic communication system of elephants is a fundamental aspect of their ecology and behavior.

Audio-based monitoring has emerged as a promising tool in elephant conservation efforts. By enabling tools such as elephant presence detection and localization, or individual recognition by vocal fingerprint, AI can offer valuable insights into elephant behavior, communication, and movement patterns. 

\subsection{Passive Acoustic Monitoring} 

Bio-acoustic data gathered from a strategically placed grid of microphones can be employed to detect the presence of elephants over a significantly wider range compared to camera traps. This approach, termed Passive Acoustic Monitoring (PAM), possesses distinct advantages over imaging techniques. Whereas imaging is constrained by directionality and is ineffective in capturing occluded subjects, PAM can identify acoustic signals encompassing a radius surrounding the microphone, though obstructions such as dense foliage. This renders PAM particularly advantageous in forested terrains where visibility is compromised. Moreover, the low frequency of infrasonic elephant rumbles can be detected over 3 km from the source \cite{Mortimer2018, Thompson2010, Hedwig2018}, facilitating a large potential radius of measurement. However, this radius is strongly influenced by factors such as foliage density, source amplitude and ambient noise levels. 

PAM presents technical challenges, as working with real-world bio-acoustic data is complex due to background noise interfering with event detection and species classification. Moreover, separating individual calls from numerous simultaneous calls, often referred to as the "cocktail party problem" in human audio analysis, is a crucial aspect of determining the number of elephants present at any given moment and it remains a formidable task. Another limitation of PAM is that elephants can only be detected when they are producing sound, which may result in some elephants going undetected.

Recent research from The Elephant Listening Project at Cornell University \cite{Wrege2017,Bjorck2019,Sethi2020,Keen2017} has explored the application of AI to PAM for elephant call detection in the Congo Basin. In one study, the detector had reasonable performance (0.8 recall) in identifying when an elephant rumble occurred in a continuous wild recording \cite{Wrege2017}, while another study achieved very similar performance (82\% true positive score) on identifying when elephant calls occured in audio recordings \cite{Keen2017}. Furthermore, these techniques have been implemented on a portable embedded system, enabling on-board AI audio processing and the subsequent transmission of results \cite{Schwartz2021}, thus creating a low-power monitoring system for potential long-term measurements in the field. Other notable works on elephant detection \cite{Wijayakulasooriya2011} used a hidden Markov model to detect elephants from continuous infrasonic rumble recordings with 97.6\% accuracy in the presence of noise, and Venter's work on elephant rumble detection \cite{Venter2009}, using a Voice Activity Decode, which achieved 90.5\% detection accuracy. 
               
For future work on elephant presence detection via PAM, data preprocessing methods can be applied to enhance species classification performance. More robust denoising techniques, for instance, help eliminate non-critical events such as weather, making it easier to detect events of interest. Such techniques have demonstrated improved performance in the classification of whale clicks \cite{Bermant2019} and songs \cite{Allen2021}. Additionally, when an elephant call is detected, multiple calls are sometimes present simultaneously. To potentially utilize PAM for census purposes \cite{Wrege2017}, it is essential to determine the number of elephants present during an elephant call event. Deep learning source separation techniques can filter the data to provide separate audio streams for calls from different sources, showing promise across various bioacoustic datasets \cite{Bermant2021}. Unsupervised source separation techniques have also demonstrated success in classification improvement \cite{Wisdom2020}, and unsupervised techniques have also been used to improve event detection \cite{Bermant2022}. The benefits of unsupervised learning are discussed more in section \ref{unsupervised}.

Not only can this technology be used to detect elephants, but it can also aid elephant conservation by detecting gunshots and chainsaws, which are indicative of potential threats to wildlife and their habitats. There exists some work in this field \cite{Van2013}, one using convolutional neural networks \cite{Katsis2022,Hrabina2015}, however, the current accuracies of these algorithms make it impractical to rely on these methods alone. One notable exception is the work done by Wrege et al., which has quite impressive performance on gunshot detection (0.94 recall) \cite{Wrege2017}. 

\subsection{Individual Identification}
Not only is there strong evidence that elephants can identify other individuals from their calls \cite{Mccomb2003,Mccomb2000}, researchers have also had some success in individual classification from audio data as well. Clemins et al. \cite{Clemins2005} demonstrated 83\% accuracy in identifying individuals by employing a hidden Markov model with feature extraction. Interestingly, more advanced AI techniques have successfully identified individual lions from their roar recordings, achieving impressive accuracy of up to 98\% \cite{Wijers2021,Trapanotto2022}. These advanced techniques could easily be adapted to elephant data, likely enhancing the accuracy of individual identification from audio. Because this work depends on training on a known population with audio data labelled with the individual caller, it is best suited for monitoring in areas where the home population has sufficient data.

\subsection{Full-Spectrum Ensemble Audio Monitoring}\label{multi-audio}
AI applications in passive acoustic monitoring tend to concentrate on specific call types or frequencies, such as rumbles or trumpets. To improve detection of elephants near monitoring stations, it's beneficial to integrate a broad set of elephant vocalizations and frequencies into a single classification system. However, given the challenges in estimating call production rates, this method is better suited for detecting elephant presence rather than accurate population censusing.

\section{Seismic Monitoring}
Seismic signals, the transmission of low-frequency vibrations through the ground, constitutes a potential component of elephant communication with significant implications for conservation efforts. As previously discussed, elephants are known to produce low-frequency audio vocalizations, known as infrasound, that can travel long distances \cite{Mortimer2018}, up to approximately 3 km \cite{Thompson2010}. Concurrently, the emission of these infrasonic calls generates seismic waves that propagate through the ground \cite{Gunther2004}. Elephants have shown to be able to detect these seismic vibrations through bone conduction and specialized mechano-receptors in their feet \cite{Bouley2007}, potentially allowing them to communicate with one another over vast expanses of their habitat, conveying information about potential threats, herd movements, resource utilization, and reproductive status \cite{OConnellRodwell2007,Reinwald2021}. Moreover, elephants can differentiate the same seismic call type from different individual elephants, owing to their ability to discriminate frequency changes within a narrow bandwidth in the low-frequency spectrum \cite{OConnellRodwell2007}.  

Seismic monitoring presents several compelling opportunities in the field of conservation. First, it provides an innovative way to detect elephant presence, which could be highly advantageous for potential future applications such as population censusing, monitoring, corridor planning and early warning systems to prevent human-elephant conflict (HEC). Second, a deeper understanding of the social behavior surrounding these seismic signals can be gained with long-distance monitoring techniques that are minimally disruptive to the animal's environment. Finally, considering elephants' ability to discern caller identity from the same call, it is plausible that additional information on elephants can be extracted with the analysis of seismic data. This may potentially include details such as the size of the caller, or the caller's identity.

\subsection{Elephant Detection}
While initial studies have primarily focused on analyzing elephant behavior and responses to these seismic signals through geophone measurement \cite{OConnellRodwell2007}, seismic monitoring has been proposed \cite{Reinwald2021,Wood2005,Anni2015} and implemented \cite{Parihar2022} as a method for elephant population monitoring, encompassing both censusing and tracking. Non-machine learning techniques have been employed to detect elephant presence from seismic data for censusing purposes, achieving 85\% accuracy \cite{wood2005_2} and, continuous wavelet transforms reached 90\% accuracy in detecting forest elephants \cite{Parihar2021}. Within the realm of machine learning, elephant calls have been classified from seismic measurements using Support Vector Machines (SVMs) with 73\% accuracy \cite{Parihar2022}, neural networks with 87\% accuracy \cite{Fernando2020}, and convolutional neural networks attaining 80-90\% accuracy up to 100m away \cite{Szenicer2022}.

Utilising AI techniques for species classification and localisation within seismic data holds promise for enhancing the accuracy of elephant censusing. As highlighted in the audio section \ref{multi-audio}, elephant infrasonic calls are accompanied by corresponding seismic calls. Each of these transmission modalities has its own mode of transport. Depending on environmental factors, either the audio or seismic signal may propagate more effectively \cite{Mortimer2017}. By integrating techniques for elephant detection across both audio and seismic modalities, it's possible to reduce noise and more accurately distinguish elephant sounds from background interference. This is particularly beneficial when the signal-to-noise level is low in both measurements, a scenario frequently encountered at extensive measurement radii \cite{Gunther2004, OConnellRodwell2007}. Such bi-modal detection can not only extend the effective detection range beyond that achievable by either audio or seismic monitoring alone but also elevate the signal-to-noise ratio (SNR) without compromising the measurement radius. This advantage becomes especially pronounced in forested or densely vegetated regions where acoustic waves face substantial attenuation \cite{Gunther2004}.

Localization of elephants is another essential task. Recent work has compared seismic and acoustic localization \cite{Reinwald2021}, revealing improved localization using seismic data. Moreover, the multimodal combination of seismic and acoustic measurements can provide location information with a limited array, as seismic and acoustic waves travel at different speeds, allowing distance to be determined by the delay between the two signals if the soil composition is known \cite{OconnellRodwell2000}.

In the context of behavior monitoring, there is considerable potential for analyzing call types and long-distance elephant communication \cite{Mortimer2018}, even though much remains uncharted. O'Connell Rodwell's research group has significantly contributed to identifying the objectives of certain calls, such as when elephants are leaving a resource or are in estrus \cite{OConnellRodwell2007}. Other researchers have effectively automated the classification of distinct behaviors through seismic measurements, including walking, running, and rumbles \cite{Mortimer2018}. The application of AI in behavior monitoring within this domain continues to offer substantial opportunities for further exploration and advancements. 

\subsection{Future of Seismic and Acoustic Monitoring} 
Lastly, the potential exists for acquiring additional information from seismic or low-frequency sound data through frequency analysis. Studies in whale bioacoustics suggest that individual characteristics, such as body size, weight, age, and sex, can be inferred through the analysis of whale songs, as the caller's unique size and vocal cord properties generate distinctive sounds associated with various phenotypes and traits \cite{Taylor2010}. Given that elephants can discern between familiar and unfamiliar callers based on their seismic signals, and that the frequency characteristics of the seismic wave an individual can transmit are influenced by their body size \cite{OConnellRodwell2007}, it is plausible that an individual's traits can be deduced from their seismic call. These traits may encompass body size, individual identity, sex, and age. Further research into the application of AI in this area will prove immensely beneficial for monitoring efforts.

\section{Olfactory Monitoring}
Olfaction, or the sense of smell, plays a critical role in the lives of elephants as well as in conservation efforts. Elephants have an extraordinary sense of smell, which they rely on for various purposes, such as foraging, locating water sources \cite{Wood2022}, identifying herd members \cite{Bates2008}, communicating with other elephants \cite{Rasmussen1998}, and navigation \cite{Allen2021_2}. The elephant's olfactory system is highly developed, featuring a large number of olfactory receptor genes and a sophisticated vomeronasal organ \cite{Rasmussen1996,Cave2019}. These adaptations enable elephants to detect and interpret a wide range of chemical signals in their environment. By studying these chemical signals, researchers can better understand elephant behaviour, social structure, and reproductive strategies. As our understanding of elephant olfaction and as measurement technology improves, new opportunities for monitoring chemical signals and how it affects elephant behaviour will arise. 

This section explores how AI analysis of olfactory measurement during elephant behavior studies can be utilized for understanding elephant social interactions and decision making. The implications of this work are large and can lead to very useful insights such as mitigating human-elephant conflicts by identifying cheap and effective odour deterrents. However, the authors urge caution in applying these findings until a thorough understanding of olfactory sensing in elephants is obtained. This will ensure the appropriate use of odor deterrents and prevent any unintended consequences on the behaviour and welfare of elephants.

Traditionally, olfactory measurements are difficult to make in the field due to hardware constraints. Current technology makes it difficult to measure the presence of a known particulate in the air. The literature surveyed suggests  that most studies on elephant olfaction either expose elephants to a known chemical, assessing their reactions, or present the same scenario to elephants both with and without the chemical \cite{Allen2021_2,Wood2022,Bates2008}. This means that in order to conduct this work, the researcher must first identify a particular chemical of interest, and then test it, which is a very costly method of research. Incorporating AI into molecular sensing technology can provide researchers with a more extensive dataset for understanding elephant behaviour. 

\subsection{AI Enhanced Chemosensors}
An electronic nose (e-Nose) or chemosensor is a device designed to mimic the biological olfactory system, capable of detecting and identifying various chemical compounds and odors in the surrounding environment. Though there exists technology to mimic the actual biological sensors in noses \cite{Cave2019}, discerning meaning from these sensors is quite difficult. Recent work has applied AI in interpreting measurements from these chemosensors to infer chemicals present in the air with reasonable results \cite{Melendez2022,Gardner1990,Fu2007}, although most work focuses on inanimate sensors based on metal-oxides or polymers and therefore are limited in what they are able to detect. Future work in applying machine learning techniques to data from biological chemosensors, sensors where the sensing mechanism is based on the same biological materials as mammalian noses \cite{Cave2019}, shows a lot of promise. Google Inc has also released a new startup researching AI for olfaction sensing \cite{google_cloud_blog_2023}. The technology from the research done in this field will lead to more portable systems, such as those seen in \cite{Melendez2022}, with higher specificity to air particulate identification. 

This increased precision in detecting airborne particulates will enable researchers to delve deeper into the role of olfactory cues in influencing elephant decisions, and help to understand elephant responses. One potential application of such high precision measurements is discerning the influence of olfactory cues on elephants' migratory patterns. While elephants have demonstrated the ability to detect water through scent \cite{Wood2022}, other species are known to utilise chemical cues for navigation \cite{Papi1989}. Improved olfactory sensing using AI can provide a tool to understanding if elephants also rely on similar cues for their migratory decisions.

\subsection{Olfactory Deterrents for Human-Elephant Conflict Mitigation}
Once certain chemicals are identified, these can be used in elephant conservation. Elephants have an aversion to certain smells, which can be utilized as olfactory deterrents to minimize human-elephant conflicts (HECs). Some deterrents, such as bees and chili peppers, have been used to create barriers around crops to deter elephants from raiding \cite{Davies2011,King2007,King2009}. These methods can help reduce crop damage and protect both humans and elephants from potentially dangerous encounters. However, some elephants may habituate to such deterrents over time, reducing their effectiveness. Moreover, the success of these deterrents can be influenced by factors such as the local environment, weather conditions, and the availability of alternative food sources for elephants. Research into understanding the elephants olfactory universe can lead to far more subtle ways to reduce human elephant conflicts.

\section{Discussion} 
In this paper, we have reviewed various AI and Machine Learning methods that can benefit elephant monitoring and conservation. The key challenge for future work in this field is fostering a consistent and long-lasting collaboration between the fields of conservation and AI. The pace at which each field advances is quite different: AI is rapidly evolving, with new advancements emerging frequently; while elephant studies require prolonged periods to establish trust with subjects, observe them, and adapt experiments to suit the population and environmental constraints. Collaborative efforts should not be limited to merely gathering data and applying AI tools. Instead, they should involve cyclical processes of data exploration, transformation, analysis, interpretation, and communication. From a conservation perspective, this requires clear goal-setting, statements of the research objectives and keeping up to date on technological advancements relevant to their work. For AI professionals, early involvement in discussions on data collection is crucial, as well as gaining an understanding of the practical challenges and timelines associated with field work.

From the discussions in this review, it is evident that AI holds substantial potential for applications in elephant research and monitoring. These applications span from data collection and curation to deriving meaningful representations of data and generating novel scientific hypotheses. However, it is important to clarify the specific contexts where AI excels and its inherent limitations. Typically, AI and ML methods thrive in scenarios with extensive datasets and well-defined research questions. Their core strength resides in pattern recognition. For instance, supervised learning targets predefined patterns that researchers aim to detect, while unsupervised learning endeavours to understand the inherent structure, pattern or characteristics of a dataset. Additionally, AI is particularly efficient in scenarios requiring consistent and repetitive computation, especially when working with standardised or normalised datasets. In contrast, AI may not be the ideal solution for more ambiguous or heterogeneous data situations.

\subsection{Transfer Learning}
In numerous video and imaging classification tasks for elephant recognition, the developed models often face difficulties in transferring to new environments due to varying environmental background features and the presence of unfamiliar species \cite{Beery2019,Beery2018}. Adapting these networks for new environments could be expensive, as it necessitates collecting and annotating new data. Several approaches can be employed to enhance a model's transfer-ability without requiring the direct measurement of new datasets in the field. First, AI model developers may utilise simulated data to broaden the training dataset, which has demonstrated promising results \cite{Beery2020}. Another strategy involves creating more general models that exhibit robustness across multiple environments, followed by fine-tuning for specific tasks of interest \cite{Beery2019}. A notable example of this is Megadetector, a tool used to detect animals in camera trap images with an impressive ability to transfer to new environments easily \cite{Beery_MD}. Finally, self-supervised or unsupervised methods can be employed for training models, enabling dataset expansion without the need for labels. 

\subsection{Self and Unsupervised Methods}\label{unsupervised} 
A significant portion of AI work in the field relies on the collection of vast amounts of data. Furthermore, the majority of techniques reviewed in this paper utilise supervised training, requiring extensive data annotation for effective model training. The combined requirements of data collection and labour-intensive annotation lead to considerable development costs for these tools. In a resource-constrained field, such as conservation, these costs pose substantial obstacles to the ongoing development and application of AI solutions.In a resource-constrained field, such as conservation, these costs pose substantial obstacles to the ongoing development and application of AI solutions. 

Self-supervised and unsupervised learning methods are AI techniques that do not rely on labelled data, allowing them to automatically discover patterns and structure within datasets without needing humans to mark or annotate the data first. This area of research is currently experiencing significant activity. Notable examples include self-supervision in images to learn valuable information from the images and videos automatically \cite{Caron2021,Grill2020}, and using unsupervised methods to improve and automate bioacoustics analysis of PAM datasets without labeling \cite{Wisdom2020,Bermant2022}. Moreover, innovative techniques have emerged for animal behavior monitoring without the need for labels using unsupervised techniques \cite{Luxem2022,Wiltschko2015}. As these methods continue to advance, they will not only reduce the costs associated with data collection but also facilitate novel insights into animal behavior that may elude human consideration.

\section{Conclusion}
Artificial intelligence techniques have emerged as powerful tools for enhancing elephant monitoring and conservation efforts. AI-based techniques in imaging, video, audio, seismic, and olfactory modalities have shown promising advancements, offering more sophisticated and efficient alternatives to traditional monitoring methods. These monitoring techniques have the potential to improve the accuracy and efficiency of elephant behaviour studies, automate data labelling and analysis, and detect nuanced behaviours or patterns that may enhance capabilities of human observers. Furthermore, the ongoing development of novel data capture hardware systems, such as drone monitoring and olfactory measurement, generates a vast array of data that can be harnessed to create powerful AI analysis tools. 

Advancing these AI models requires consistent collaboration between AI specialists and conservationists and some methods, tools and analyses will need to be uniquely tailored for elephant research. These tools will arise primarily from needs within that field, for example, identifying individual elephants. For this work input from elephant experts is essential, and relying solely on adjusting general models will not be adequate for these specialised needs. 

Additionally, challenges remain in transferring AI models to new environments, considering the high cost of data collection and annotation and the limitations of supervised learning techniques. Future research should focus on strategies such as using simulated data, fine-tuning pretrained models, and employing self-supervised or unsupervised techniques to address these data shortcomings and promote wider adoption of AI in elephant conservation efforts.

Finally, while each monitoring modality has been addressed individually in this overview, integrating various modalities in future work can provide a comprehensive understanding of the stimuli elephants encounter and their perception of their environment, and can aid in understanding how elephants make decisions.

AI offers significant advancements in animal monitoring, with elephants being particularly apt for the application of these techniques. As AI applications in this domain evolve and challenges are addressed, our understanding and protection of elephants will improve. This not only supports the longevity of elephants and their habitats but also paves the way for technologies that could aid broader conservation initiatives.

\section*{Acknowledgments} 
We would like to thank Matt James for his extensive knowledge and guidance in the field of elephant conservation. His direction was invaluable in connecting us with the right experts, which significantly informed our work and understanding. 
We also wish to acknowledge the use of AI language services which contributed to the editing and improvement of the readability of this manuscript. As befits the subject, an AI service was utilized for its editing, while the content of the paper was independently generated by the only too human authors.

\section*{Funding} 
This work was internally funded by Colossal Laboratories \& Biosciences.



\bibliographystyle{unsrt}  
\bibliography{library}

\end{document}